\documentclass[11pt]{article}

\usepackage{arxiv}

\usepackage{amsmath}
\usepackage{amssymb}
\usepackage{array}
\usepackage{booktabs}
\usepackage{graphicx}
\usepackage{multirow}
\usepackage{natbib}
\usepackage{hyperref}
\usepackage{longtable}
\usepackage{xcolor}
\usepackage{colortbl}
\usepackage{tikz}
\usetikzlibrary{arrows.meta,positioning,shapes.geometric,fit,backgrounds}

\hypersetup{
  colorlinks=true,
  linkcolor=blue,
  citecolor=blue,
  urlcolor=blue
}

\usepackage[ruled,vlined,linesnumbered]{algorithm2e}
\usepackage{float}
\usepackage{wrapfig}
\usepackage{caption}
\floatname{algorithm}{Algorithm}
\SetAlgoCaptionSeparator{:}
\SetAlCapFnt{\bfseries}
\SetAlCapNameFnt{\bfseries}
\SetKwInput{KwIn}{Input}
\SetKwInput{KwOut}{Output}
\SetNlSty{textbf}{}{}
\SetAlgoNlRelativeSize{-1}
\IncMargin{1.75em}

\newcommand{\offload}{\ensuremath{\tau_{\mathrm{off}}}}
\newcommand{\localctx}{\ensuremath{\mathcal{C}_s}}
\DeclareRobustCommand{\flowstep}[1]{%
  \raisebox{-0.12ex}{\tikz[baseline=(c.base)]{%
    \node[circle,draw,line width=0.35pt,inner sep=0.45pt,minimum size=1.05em,font=\scriptsize] (c) {#1};%
  }}%
}

\date{}

\usepackage{marvosym}
\usepackage{XCharter}
\usepackage[xcharter,bigdelims,vvarbb]{newtxmath}

\usepackage{enumitem}
\usepackage{tcolorbox}
\tcbuselibrary{breakable,skins}

\makeatletter
\newcommand{\normpyromind}{%
  \@setfontsize\normpyromind{10pt}{11pt}%
  \abovedisplayskip 9pt \@plus 2pt \@minus 3pt
  \abovedisplayshortskip 4pt \@plus 2pt \@minus 2pt
  \belowdisplayskip 9pt \@plus 2pt \@minus 3pt
  \belowdisplayshortskip 6pt \@plus 2pt \@minus 2pt
}
\renewcommand{\normalsize}{\normpyromind}
\makeatother
\DeclareMathSizes{10}{11}{8}{6}
\normpyromind
\definecolor{PyroInk}{HTML}{171B2C}
\definecolor{PyroIndigo}{HTML}{35339A}
\definecolor{PyroViolet}{HTML}{7B3FF2}
\definecolor{PyroGray}{HTML}{8A8A8A}

\tcbset{
  pyrocasebox/.style={
    enhanced,
    breakable,
    boxrule=0.55pt,
    arc=2pt,
    outer arc=2pt,
    left=8pt,
    right=8pt,
    top=5pt,
    bottom=5pt,
    before skip=5pt,
    after skip=6pt,
    fonttitle=\normalfont\bfseries\sffamily\small,
    fontupper=\small,
    before upper={%
      \setlength{\abovedisplayskip}{5pt}%
      \setlength{\belowdisplayskip}{5pt}%
      \setlength{\abovedisplayshortskip}{3pt}%
      \setlength{\belowdisplayshortskip}{3pt}%
    },
    before lower={%
      \setlength{\abovedisplayskip}{5pt}%
      \setlength{\belowdisplayskip}{5pt}%
      \setlength{\abovedisplayshortskip}{3pt}%
      \setlength{\belowdisplayshortskip}{3pt}%
    }
  }
}
\newtcolorbox{pyropromptbox}[1][User prompt]{
  pyrocasebox,
  title={#1},
  colframe=PyroIndigo!80!black,
  colback=PyroIndigo!3,
  colbacktitle=PyroIndigo!88!black,
  coltitle=white
}
\newtcolorbox{pyrotracebox}[1]{
  pyrocasebox,
  title={#1},
  colframe=PyroGray!70!PyroInk,
  colback=PyroIndigo!2,
  colbacklower=PyroViolet!2,
  colbacktitle=PyroInk!7,
  coltitle=PyroInk,
  segmentation style={solid,draw=PyroGray!60,line width=0.45pt},
  fontlower=\small
}
\newcommand{\pyroboxlabel}[1]{%
  \noindent{\normalfont\bfseries\sffamily\footnotesize
  \color{PyroIndigo}#1}\par\smallskip
}

\hypersetup{
  pdftitle={PyroDash: Cost-Efficient Token-Level Small-Large Model Collaborative Inference},
  pdfauthor={PyroMind Dynamics},
  pdfsubject={Research preprint},
  pdfkeywords={SLM-LLM collaboration, introspective routing, GRPO, cost-efficient inference, collaborative intelligence},
  linkcolor=PyroIndigo,
  citecolor=PyroIndigo,
  urlcolor=PyroViolet
}

\AtBeginDocument{%
  \newgeometry{%
    left=2.2cm,
    right=2.2cm,
    top=3cm,
    bottom=3cm,
    headheight=40pt,
    headsep=20pt,
    footskip=30pt
  }%
}

\setlist[itemize]{noitemsep,topsep=0.35\baselineskip}
\setlist[enumerate]{noitemsep,topsep=0.35\baselineskip}

\newcommand{\pyromindbrand}{%
  \includegraphics[
    height=18pt,
    trim=44.3bp 26bp 29bp 47.3bp,
    clip
  ]{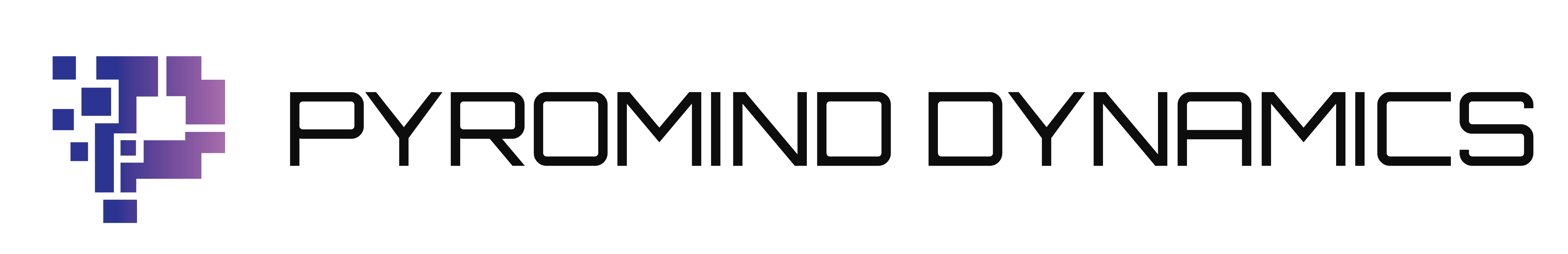}%
}

\newcommand{\pyroheaderfont}{%
  \normalfont\fontfamily{lmss}\fontsize{8}{10}\selectfont
}
\newcommand{\pyrofooterfont}{%
  \normalfont\fontfamily{lmss}\fontsize{8}{10}\selectfont\color{black}
}

\fancypagestyle{pyromindfirst}{%
  \fancyhf{}%
  \fancyhead[L]{\pyromindbrand}%
  \fancyhead[R]{%
    \pyroheaderfont
    \href{https://pyromind.ai/}{pyromind.ai}\\
    \itshape Research Preprint%
  }%
  \fancyfoot[L]{%
    \pyrofooterfont\itshape
    \Letter~Corresponding author(s): \href{mailto:kevin.ding@pyromind.ai}{%
      \textcolor{black}{kevin.ding@pyromind.ai}%
    }%
  }%
}

\makeatletter
\let\PyroDefaultHeadRule\headrule
\let\PyroDefaultFootRule\footrule
\renewcommand{\headrule}{\color{PyroGray}\PyroDefaultHeadRule}
\renewcommand{\footrule}{\color{PyroGray}\PyroDefaultFootRule}

\renewcommand{\maketitle}{%
  \par
  \begingroup
    \setlength{\parindent}{0pt}%
    \begin{flushleft}
      {\raggedright\normalfont\bfseries\fontsize{21}{23}\selectfont
       \color{black}\@title\par}%
      \vskip 11pt
      {\raggedright\normalfont\bfseries\fontsize{9}{13}\selectfont
       \@author\par}%
      \vskip 20pt
    \end{flushleft}
  \endgroup
  \thispagestyle{pyromindfirst}%
}
\makeatother

\renewenvironment{abstract}
  {\par\noindent\bfseries\fontsize{10}{12}\selectfont\ignorespaces}
  {\par\vskip 1em}

\renewcommand{\keywords}[1]{%
  \par\noindent{\itshape\fontsize{9}{11}\selectfont
    \def\and{\unskip\nobreak\enspace$\cdot$\enspace\ignorespaces}%
    Keywords: #1}%
  \par\vskip 1.2em
}

\newcommand{\pyroheadingfont}{%
  \normalfont\bfseries\fontsize{13}{14}\selectfont\color{black}%
}
\makeatletter
\renewcommand{\@seccntformat}[1]{%
  \csname the#1\endcsname.\hspace{0.5em}%
}
\renewcommand{\section}{%
  \@startsection{section}{1}{\z@}%
  {-3ex \@plus -4pt \@minus -3pt}%
  {5pt}%
  {\pyroheadingfont}%
}
\renewcommand{\subsection}{%
  \@startsection{subsection}{2}{\z@}%
  {-2.5ex \@plus -3pt \@minus -2pt}%
  {2pt}%
  {\normalfont\bfseries\color{black}}%
}
\renewcommand{\subsubsection}{%
  \@startsection{subsubsection}{3}{\z@}%
  {-2ex \@plus -2.5pt \@minus -1.5pt}%
  {2pt}%
  {\normalfont\bfseries\itshape\color{black}}%
}
\makeatother

\DeclareCaptionLabelSeparator{pyropipe}{\enspace$\vert$\enspace}
\title{PyroDash: Cost-Efficient Token-Level Small-Large Language Model Collaborative Inference}

\renewcommand{\undertitle}{}
\renewcommand{\headeright}{}
\renewcommand{\shorttitle}{PyroDash: Cost-Efficient Token-Level Small-Large Model Collaborative Inference}

\author{
  Niqi Lyu, Pengtao Shi, Wei Qiu, Jianlin Zhong, Sicong Xia, Jianyao Ma, Yicheng Ding\textsuperscript{~\Letter}  \\
  {\normalfont\fontsize{8}{10}\selectfont Pyromind Dynamics Inc.} \\
}

\begin{document}

\maketitle

\begin{abstract}
Large language models (LLMs) provide strong reasoning capabilities but are expensive to serve at scale, whereas small language models (SLMs) are cheaper but less reliable on difficult problems. We introduce PyroDash, a cost-aware framework for token-level SLM-LLM collaborative inference. During generation, the SLM decides whether to request assistance by emitting a control token. A Collaborate Engine then sends the query and partial reasoning trace to a frozen LLM for completion through a single handoff. The policy is internalized in the SLM, requiring neither a separate router, LLM retraining, nor access to LLM logits. PyroDash trains the SLM in three stages: control-token embedding learning, offloading-oriented supervised fine-tuning, and cost-aware alignment with Group Relative Policy Optimization. Its reward balances answer accuracy against inference cost normalized by LLM-only inference. Across five mathematical reasoning benchmarks, PyroDash supports different accuracy-cost operating points. With $\lambda=0.05$, it achieves 64.04 percent average accuracy, 6.36 percentage points above the LLM-only baseline, while reducing cost by 20.4 percent. With $\lambda=0.6$, it achieves 54.55 percent accuracy with a 1.90 percent LLM token ratio and 0.012 LLM calls per example, reducing total cost from USD 49.36 to USD 1.78. These results show that learned token-level handoffs can reduce LLM use while preserving strong reasoning performance.
\end{abstract}

\keywords{SLM--LLM Collaboration \and Token-Level Self-Routing \and GRPO \and Inference Cost \and Collaborative Inference}

\section{Introduction}

Large language models (LLMs) have demonstrated strong capabilities in logical reasoning, mathematical problem solving, and symbolic computation~\citep{achiam2023gpt4,touvron2023llama2,yang2024qwen2}.
Their real-world deployment, however, must balance multiple requirements, including reasoning quality, interaction latency, privacy compliance, and inference cost~\citep{chen2026networkedge,li2025edgeslm}.
As agentic workflows, long-context interactions, and multi-step reasoning become increasingly common, making inference efficient has become as important as increasing model capacity.
Cloud-hosted LLMs provide access to centralized compute and strong model capabilities.
Yet serving every request with an LLM can produce substantial API bills, particularly in high-frequency or long-generation workloads.
Transmitting user inputs to remote servers can also raise privacy and data-governance concerns~\citep{chen2026networkedge,li2025edgeslm}.
In contrast, open-weight sub-10B variants from model families such as Gemma~4~\citep{gemmateam2026gemma4} and Qwen3.5~\citep{qwenteam2026qwen35} make low-cost, self-hosted inference increasingly practical.
Their smaller computational cost, however, comes with weaker generalization and a limited ability to solve problems that require difficult or extended reasoning.
Quantization~\citep{frantar2023gptq} and distillation~\citep{gu2024minillm} can further improve these compact models under a fixed resource budget, but compression alone cannot reliably close the capability gap with LLMs.
Consequently, neither model scale is sufficient on its own: SLMs are economical but may fail on hard problems, whereas LLMs are capable but costly to invoke indiscriminately.
Collaborative inference provides a promising way to address this dilemma by combining the complementary strengths of small and large language models~\citep{ding2024hybridllm,shen2024collm,zheng2025citer,li2025edgeslm}.
Instead of routing each request entirely to the SLM or the LLM, we allow the SLM to reason independently and request LLM assistance only when a reasoning step exceeds its capabilities.
The central challenge then shifts from \emph{which model to deploy} to \emph{when, during reasoning, the stronger model should intervene}.

Existing methods address only part of this challenge.
Request-level routing and cascaded serving choose a model before decoding begins~\citep{chen2023frugalgpt,ong2024routellm,ding2024hybridllm}.
Although effective for queries whose difficulty is evident from the input, they cannot react when a seemingly simple problem becomes difficult only at an intermediate reasoning step.
Token-level collaboration moves the decision into the generation process~\citep{zheng2025citer,shen2024collm,huang2026relayllm}, but existing designs may require a separate router, repeated switching between models, or access to the target LLM's internal outputs.
Speculative decoding similarly coordinates small and large models, yet primarily optimizes latency or throughput and often assumes access to target-model logits or weights~\citep{leviathan2023fast,chen2023accelerating,xia2024unlocking}.
Most importantly, these approaches rarely train the routing policy against the billed inference cost incurred under practical serving prices.
Together, these limitations leave two questions unresolved.
(i) \textit{Can an SLM recognize, from its own generation trajectory, when stronger assistance is needed?}
(ii) \textit{Can it learn this decision under an explicit objective that balances answer accuracy against inference cost?}
A practical solution should answer both questions while avoiding repeated handoffs, keeping the LLM frozen, and remaining compatible with proprietary APIs.
The cost-aware handoff timing is particularly important because commercial APIs commonly price input and output tokens separately: the handoff point determines both the context-prefill cost and the remaining autoregressive-decoding cost of the LLM.

We answer the above questions with \textbf{PyroDash}, a novel cost-aware, token-level paradigm for collaborative inference between SLMs and LLMs.
During autoregressive generation, the SLM either completes the reasoning independently or emits the dedicated control token \offload{} when it predicts that stronger assistance is needed.
Upon detecting this token, the Collaborate Engine packages the original query together with the partial reasoning trace.
It then transfers this combined context to a frozen LLM, which completes the remaining reasoning in a single handoff.
This division of responsibility places the routing policy within the SLM and limits the Collaborate Engine to executing the handoff protocol.
As a result, PyroDash requires no separate learned router.
Because the LLM serves only as a frozen continuation model, PyroDash also requires neither LLM retraining nor access to its internal logits, making it compatible with proprietary LLM APIs.

Training the SLM to decide when to reason independently and when to offload requires both a reliable representation of the new control token and an offloading policy aligned with deployment cost.
To meet these requirements, PyroDash uses a three-stage progressive training pipeline for the SLM: control-token embedding learning, offloading-oriented supervised fine-tuning for behavioral cold start, and cost-aware policy alignment with Group Relative Policy Optimization (GRPO)~\citep{shao2024deepseekmath}.
The first stage integrates \offload{} into the SLM's representation space, and the second teaches the model both to reason independently and to request assistance conditionally.
In the final stage, GRPO updates the SLM's offloading policy using a reward that combines answer accuracy with an inference-cost penalty normalized by the cost of LLM-only inference.
Varying the penalty coefficient $\lambda$ yields a family of offloading policies for different accuracy--cost preferences.

We consider mathematical reasoning tasks a natural testbed for adaptive offloading because they vary widely in difficulty: easier problems test the SLM's ability to reason independently, whereas harder problems test its ability to recognize when LLM assistance is needed.
Thus, we run our evaluation on five benchmarks---GSM8K~\citep{cobbe2021gsm8k}, Minerva~\citep{lewkowycz2022minerva}, OlympiadBench~\citep{he2024olympiadbench}, AIME25~\citep{lin2025aime}, and AIME24~\citep{huggingfaceh4_2024_aime}.
When accuracy is prioritized ($\lambda{=}0.05$), PyroDash reaches an average accuracy of 64.04\%, exceeding the LLM-only baseline's accuracy of 57.68\% by 6.36 percentage points while reducing total cost by 20.4\%.
When cost reduction is prioritized ($\lambda{=}0.6$), PyroDash achieves 54.55\% average accuracy with an LLM token ratio of 1.90\% and 0.012 LLM calls per example.
Under the evaluated pricing model, the resulting policy reduces total inference cost from \$49.36 to \$1.78, corresponding to a 96.4\% reduction.
These results show that different values of $\lambda$ yield policies with different levels of LLM reliance, enabling PyroDash to support different accuracy--cost preferences.
Beyond these empirical gains, PyroDash suggests a collaborative scaling paradigm that extends beyond increasing the size of any single model.
By organizing heterogeneous SLM and LLM services into a symbiotic mesh, compute and intelligence can be allocated on demand across model endpoints.
Such collaboration could make advanced reasoning more accessible while controlling inference cost.

Our main contributions are threefold:
\begin{itemize}
  \item \textbf{Token-level Collaborative Inference:} We introduce PyroDash, which internalizes single-handoff routing within the SLM, keeps the LLM frozen, and requires no separate learned router.
  \item \textbf{Cost-aware Alignment with RL:} We develop a three-stage pipeline to train the offloading policy of the SLM, culminating in alignment with a novel GRPO reward for tunable quality--cost control.
  \item \textbf{Substantial Inference Cost Reduction:} Across five mathematical reasoning benchmarks, PyroDash reduces inference cost by 96.4\% at $\lambda{=}0.6$, with an LLM token ratio of 1.90\% and average accuracy 3.13 percentage points below the LLM-only baseline; at $\lambda{=}0.05$, it exceeds the LLM-only baseline by 6.36 percentage points.
\end{itemize}

\section{Related Work}
\label{sec:related-work}

\textbf{Request-level routing and cascaded serving.} FrugalGPT~\citep{chen2023frugalgpt} builds API cascades; RouteLLM~\citep{ong2024routellm} and Hybrid LLM~\citep{ding2024hybridllm} route by preference or predicted difficulty; MixLLM~\citep{wang2025mixllm} and LLM Bandit~\citep{li2025llmbandit} adapt selection to model pools or user preferences.
Thresholds and cascade order allow these systems to adjust the trade-off between quality and cost, and cascades may escalate after scoring an earlier response.
Nevertheless, each invocation produces a request-level answer rather than contributing to one shared reasoning trajectory, so escalation repeats generation instead of continuing from a precise failure point.
Despite reducing average cost, they decide before decoding, cannot react to hard intermediate steps, and may commit the whole query to either model.
PyroDash instead uses the SLM's partial trajectory to offload only when assistance becomes necessary.

\textbf{Token-level collaboration.} CITER~\citep{zheng2025citer} trains a cost-aware MLP router for per-token model selection, Co-LLM~\citep{shen2024collm} learns interleaved decoding as latent model selection, and RelayLLM~\citep{huang2026relayllm} requests bounded LLM spans before returning control to the SLM.
These methods improve routing granularity, but require per-token router evaluation or permit repeated model switches.
Interleaving also requires the serving loop to coordinate both models and preserve model-specific decoding state.
With stateless APIs, the serving loop may resend a growing context after repeated interventions, turning short LLM spans into additional prefill charges that token-share or FLOP-based metrics can obscure.
Moreover, Co-LLM lacks an explicit cost objective, while RelayLLM penalizes LLM-token share rather than billed prefill and decoding cost.
PyroDash internalizes the decision in the SLM and limits each request to one handoff.

\textbf{Decoding acceleration.} Speculative decoding uses an SLM to draft blocks for parallel verification by the target LLM, reducing latency while preserving the target distribution~\citep{leviathan2023fast,chen2023accelerating,xia2024unlocking}.
Subsequent systems adapt this paradigm to concrete serving constraints: SpecInfer organizes candidate continuations into token trees for parallel verification in distributed and offloaded serving~\citep{miao2024specinfer}; Medusa replaces a separate draft model with multiple decoding heads and tree attention~\citep{cai2024medusa}; and MagicDec uses sparse-KV-cache drafting for high-throughput, long-context workloads~\citep{sadhukhan2025magicdec}.
Because the LLM scores every block, classical schemes require its output probabilities and keep it involved throughout generation; rejected drafts can also increase computation.
Their speedup depends strongly on draft acceptance and hardware parallelism, while poor alignment wastes both drafting and verification work.
Moreover, exact verification keeps the target LLM authoritative at every step, so the SLM cannot independently complete easy requests or improve the target model's answer quality.
They therefore depend on close model integration and optimize latency rather than LLM invocation or API bills.
PyroDash remains compatible with generation-only APIs and invokes the LLM only after one cost-aware trigger.

\textbf{Knowledge distillation and behavioral alignment.} Knowledge distillation strengthens an SLM by transferring an LLM's generation behavior into the smaller model before deployment.
MiniLLM~\citep{gu2024minillm} aligns the student and teacher generation distributions, while GKD~\citep{agarwal2024gkd} performs on-policy distillation with student-generated samples to reduce the mismatch between training and inference.
PyroDash uses SFT only to cold-start the offloading behavior; its subsequent GRPO stage instead learns when the SLM should retain control and when it should hand the partial trajectory to a frozen LLM, making inference-time collaboration complementary to offline distillation.

Overall, prior work addresses request-level cost, fine-grained scheduling, decoding latency, or offline capability transfer, but does not align generation with the cost for LLM inference.
PyroDash combines an SLM-internalized trigger, one handoff to a frozen LLM and a GRPO reward normalized against the LLM-only inference cost.
It learns an explicit accuracy--cost trade-off without a separate router or repeated LLM calls.

\section{Method}
\label{sec:method}

This section formalizes cost-aware, token-level SLM--LLM collaborative inference and presents the two components of PyroDash: an SLM-internalized, single-handoff architecture and a three-stage progressive training pipeline comprising control-token embedding learning, offloading-oriented supervised fine-tuning for behavioral cold start, and cost-aware policy alignment with Group Relative Policy Optimization (GRPO).

\subsection{Problem Formulation}
\label{subsec:problem-formulation}

SLM--LLM collaborative inference entails more than deploying a small and a large model within the same serving system.
It coordinates their roles during autoregressive decoding by dynamically allocating computation and control subject to practical constraints such as latency, privacy, bandwidth, and inference cost~\citep{chen2026networkedge,li2025edgeslm}.
Accordingly, a collaborative inference system can be characterized along three dimensions: \emph{when} offloading occurs, \emph{which} component initiates it, and \emph{at what granularity} collaboration operates.
PyroDash focuses on token-level, SLM-initiated collaboration: the SLM makes the offloading decision from its own generation trajectory and, when it emits \offload{}, the Collaborate Engine (CE) transfers the request to the LLM through a single handoff.

Formally, let $M_s$ denote a trainable small model and $M_l$ a frozen large model.
Let $P_s$ denote the fixed offloading prompt supplied to $M_s$, which specifies the SLM-first reasoning behavior and defines \offload{} as an available control action; let $P_l$ denote the fixed completion prompt supplied to $M_l$, which instructs it to continue from the user query and the partial trajectory produced by $M_s$ after a handoff.
Let $(q,y) \sim \mathcal{D}$ denote an example consisting of a user query $q$ and its reference answer $y$.
The system performs collaborative decoding for $q$ under an offloading policy $\pi_s$ parameterized by $M_s$.
At each decoding step, $\pi_s$ either continues generation with $M_s$ or transfers the partial trajectory to $M_l$ for completion.
This process produces a joint output $O(q;\pi_s)$ and incurs an inference cost $\mathrm{Cost}(q;\pi_s)$.

\textbf{The objective.} We seek an SLM-internalized offloading policy that completes reasoning with $M_s$ and invokes the frozen LLM through a single handoff only when the SLM generation trajectory reaches a step requiring stronger assistance.
To align this token-level decision with different accuracy--cost preferences, PyroDash maximizes expected answer accuracy while penalizing collaborative inference cost relative to LLM-only inference:
\begin{equation}
  \max_{\pi_s} \;\; \mathbb{E}_{(q,y) \sim \mathcal{D}}\!\left[
    \mathrm{Acc}\!\left(O(q;\pi_s),y\right)
    - \lambda \, \mathrm{Cost}_{\mathrm{norm}}(q;\pi_s)
  \right],
  \label{eq:problem-formulation}
\end{equation}
Here, $\mathrm{Acc}(O,y)$ equals 1 if the answer extracted from output $O$ is equivalent to the reference answer $y$, and 0 otherwise.
$\mathrm{Cost}_{\mathrm{norm}}(\cdot)$ is the collaborative inference cost normalized by the cost of executing the same query with $M_l$ alone.
The coefficient $\lambda \geq 0$ controls the accuracy--cost trade-off: larger values place greater emphasis on reducing inference cost.
The remainder of this section introduces the architecture of PyroDash, followed by the three-stage pipeline used to learn $\pi_s$.

\begin{figure}[!t]
    \begin{center}
        \scalebox{1}[0.92]{\includegraphics[width=.85\textwidth]{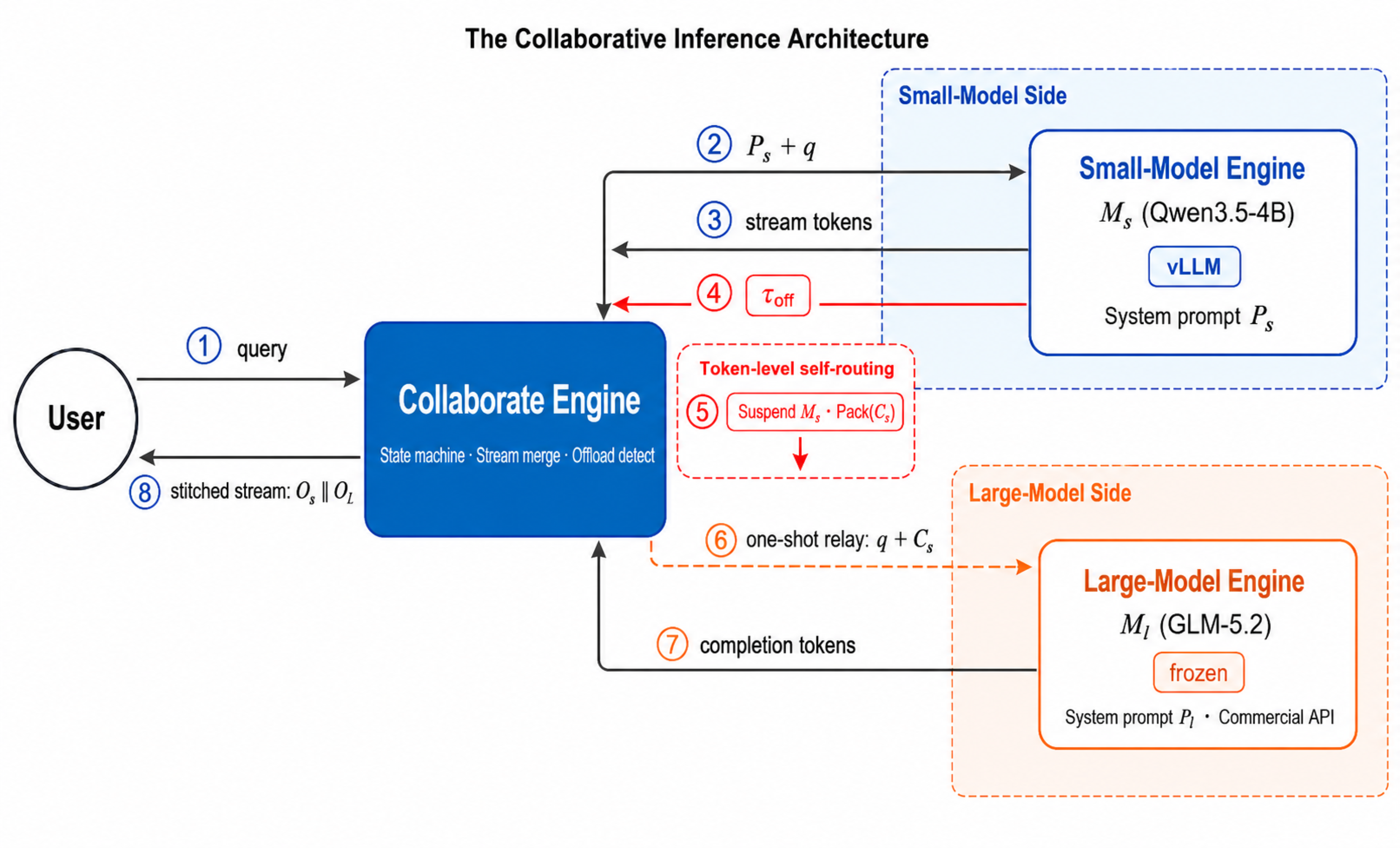}}
        \caption{The collaborative inference architecture of PyroDash (small-model-first, one-shot large-model completion). The data flow is labeled with \flowstep{1}--\flowstep{8}: \flowstep{1} $q$ enters the CE; \flowstep{2} the CE sends $P_s$ and $q$ to $M_s$; \flowstep{3} $M_s$ streams generated tokens back to the CE; \flowstep{4} \offload{} appears in the stream; \flowstep{5} the CE terminates SLM decoding and packages \localctx{}; \flowstep{6} the CE initiates a single handoff to $M_l$ using $q+\localctx{}$; \flowstep{7} $M_l$ returns completion tokens; and \flowstep{8} the CE concatenates $O_s$ and $O_l$ into the joint output $O$ and streams it to the user.}\label{fig:architecture}
    \end{center}
\end{figure}

\subsection{The Collaborative Inference Architecture of PyroDash}
\label{subsec:architecture}

Having defined the above objective, we now introduce how PyroDash executes the offloading policy $\pi_s$.
The architecture converts a token-level decision within $M_s$ into an API-compatible handoff that preserves SLM-generated reasoning without a separate router or repeated model switches.
This design comprises three components and supports two request paths, with context transferred from the SLM to the LLM at most once per request.

Figure~\ref{fig:architecture} shows PyroDash as a small-model-first decoding process.
An SLM engine hosts the trainable $M_s$, an LLM engine exposes the frozen $M_l$, and a Collaborate Engine (CE) coordinates them.
The offloading decision remains inside $M_s$; the CE only executes token-driven transitions.
For query $q$, Lines~1--2 of Algorithm~\ref{alg:ce-decode} initialize the output buffer and begin autoregressive decoding by $M_s$ conditioned on $(P_s,q)$.
The prompt $P_s$ enables \offload{} while instructing $M_s$ to attempt the solution before offloading.
If $M_s$ does not emit \offload{}, Lines~9--11 stream and accumulate the SLM-generated tokens, returning $O=O_s$ without an LLM call.
Making the offloading decision during decoding lets the policy use intermediate reasoning rather than the input alone.

Lines~3--5 of Algorithm~\ref{alg:ce-decode} specify the offloading branch.
When $M_s$ emits \offload{} at a capability boundary, the CE detects the offloading control token at Line~3, applies $\operatorname{Pack}$ to the preceding partial reasoning trace at Line~4, and terminates SLM decoding at Line~5.
The packing operation constructs \localctx{} from the SLM-generated tokens preceding \offload{}, thereby excluding the control token itself.
The LLM then uses \localctx{} as context to continue the reasoning process.
Because the same autoregressive policy generates both the reasoning trace and the offloading control token, the SLM uses its partial generation trajectory to decide when to request assistance, without relying on an external router.

Lines~6--8 of Algorithm~\ref{alg:ce-decode} specify LLM completion and joint-output construction.
At Line~6, the CE performs the single handoff by invoking $M_l$ with $(P_l,q,\localctx{})$, streams the LLM completion $O_l$ at Line~7, and constructs the joint output $O_s\mathbin{\|}O_l$ at Line~8.
The prompts have asymmetric roles: $P_s$ governs when SLM generation relinquishes control, while $P_l$ directs the frozen LLM to continue from the query and partial context to a final answer.
The control flow does not return to $M_s$, so a request ends with either SLM-only output or one-shot SLM-to-LLM completion and incurs at most one LLM API call.

\begin{algorithm}[!t]
  \footnotesize
  \caption{Single-Handoff Collaborative Decoding in the Collaborate Engine}
  \label{alg:ce-decode}
  \newcommand{\CElineNote}[2]{\makebox[0.52\linewidth][l]{Line #1: #2}}
  \KwIn{query $q$; fixed prompts $P_s,P_l$; SLM $M_s$; frozen LLM $M_l$}
  \KwOut{joint output $O$ (streamed)}
  $O_s \gets \emptyset$\tcp*[r]{\CElineNote{1}{initialize the output buffer}}
  \ForEach(\tcp*[f]{\CElineNote{2}{autoregressively decode one SLM token}}){token $t$ from $M_s(P_s,q)$}{
    \If(\tcp*[f]{\CElineNote{3}{detect the offloading control token}}){$t=\offload{}$}{
      $\localctx{}\gets\text{Pack}(O_s)$\tcp*[r]{\CElineNote{4}{construct \localctx{} without \offload{}}}
      stop decoding with $M_s$\tcp*[r]{\CElineNote{5}{terminate SLM decoding}}
      $O_l \gets M_l.\textsc{Complete}(P_l,q,\localctx{})$\tcp*[r]{\CElineNote{6}{invoke the LLM through a single handoff}}
      stream $O_l$ to user\tcp*[r]{\CElineNote{7}{stream the LLM completion}}
      \KwRet{$O_s \mathbin{\|} O_l$}\tcp*[r]{\CElineNote{8}{construct the joint output}}
    }
    stream $t$ to user\tcp*[r]{\CElineNote{9}{stream an SLM-generated token}}
    $O_s \gets O_s \mathbin{\|} t$\tcp*[r]{\CElineNote{10}{append the token to the output buffer}}
  }
  \KwRet{$O_s$}\tcp*[r]{\CElineNote{11}{return the SLM-only output}}
\end{algorithm}

Because $M_s$ expresses its offloading decision by emitting \offload{}, PyroDash can operate through standard text-generation APIs.
The LLM requires neither retraining nor logit access, while the CE needs only token detection, context packaging, and a standard completion interface.
The architecture is therefore model- and provider-independent; Section~\ref{sec:evaluation} describes the experimental deployment.

\subsection{The Three-Stage Progressive Training Pipeline for Collaborative Inference}
\label{subsec:training}

The preceding subsection introduces how the CE executes an offloading decision, but the CE itself is deliberately non-adaptive: all decisions about whether and when to hand off are made by $M_s$ within its generation stream.
The small model, however, neither represents \offload{} as a control action nor associates it with the capability boundary at which an LLM completion becomes worthwhile.
It must therefore learn to complete a query without LLM assistance when it can and to request assistance when the reasoning exceeds the SLM's capabilities.
Before handing off, it should produce a useful partial trajectory \localctx{} that can be transferred to the LLM for continuation.
This policy must also avoid handoffs whose accuracy benefit does not justify the cost.
Training $M_s$ therefore turns the deterministic procedure in Section~\ref{subsec:architecture} into the learned policy $\pi_s$ optimized by Equation~\ref{eq:problem-formulation}.

\begin{figure}[!t]
  \centering
  \includegraphics[width=0.8\textwidth]{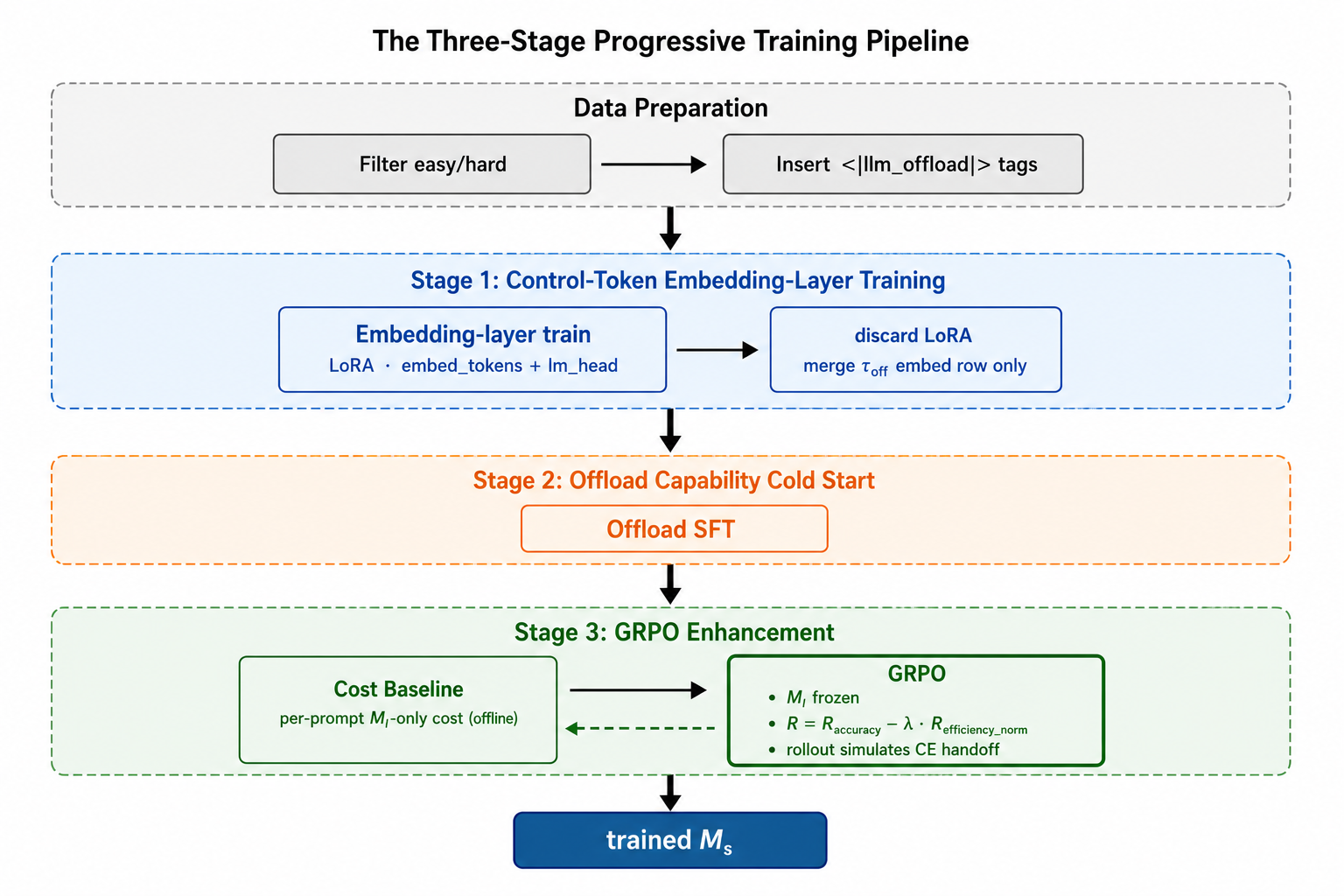}
  \caption{The three-stage progressive training pipeline. After easy/hard filtering and \offload{} label insertion, Stage~1 performs control-token embedding learning, Stage~2 applies offloading-oriented SFT for behavioral cold start, and Stage~3 performs cost-aware policy alignment with GRPO using an $M_l$-only cost baseline and handoff rollouts, producing the trained $M_s$ while $M_l$ remains frozen.}
  \label{fig:training-pipeline}
\end{figure}

PyroDash addresses this learning problem with the progressive pipeline as shown in Figure~\ref{fig:training-pipeline}.
Stage~1 performs control-token embedding learning to learn a trainable input and output representation for \offload{}.
Stage~2 uses offloading-oriented supervised fine-tuning (SFT) for behavioral cold start, establishing prompt-conditional SLM-only completion and offloading behavior.
Stage~3 performs cost-aware policy alignment with GRPO~\citep{shao2024deepseekmath}, using complete collaborative rollouts to balance answer accuracy against normalized inference cost while keeping $M_l$ frozen.
The stages thus progress from representing the handoff action to imitating the two inference paths and finally to optimizing when that action should be taken.
We next describe the data that supports these behaviors before introducing each training stage.

\subsubsection{Dataset Curation}
\label{subsec:dataset-construction}

We construct the EasyHard-24k dataset~\citep{pyromind2026easyhard24k}, which contains \textbf{24{,}061} examples in the standard \texttt{messages} format and can be loaded directly by the training framework.
The dataset is designed around three distinctions that the offloading policy must learn: whether a query is within the current capability of $M_s$, whether the collaborative mode is enabled, and where a handoff action may occur within an autoregressive trajectory.
Supervising only the presence of \offload{} would conflate these decisions and could teach the model to request assistance whenever the token is mentioned, rather than when assistance is useful.

We use the capabilities of the base $M_s$ as the reference for determining example difficulty.
Specifically, we run the base $M_s$ on each collected example and compare its response with the ground truth; examples answered correctly form the easy subset, while examples answered incorrectly but for which a correct reasoning trajectory can be reconstructed form the hard subset.
This criterion is more directly aligned with collaborative inference than a static benchmark difficulty label: an easy example demonstrates that the SLM-only path is sufficient for the current $M_s$, whereas a hard example identifies a candidate for LLM assistance.
Because the base-model trajectory of a hard example leads to an incorrect result, using that trajectory as an SFT target would reinforce the very failure that motivates offloading.
We instead ask $M_l$ to reconstruct a concise chain of thought (CoT)~\citep{wei2022chain} that leads to the ground-truth answer or \texttt{tool\_calls}.
This reconstruction supplies a valid reasoning trajectory in which candidate handoff positions can be introduced without changing the target outcome.

The easy/hard split indicates whether assistance may be useful, but it does not indicate whether the offloading action is available in the current prompting mode.
PyroDash should emit \offload{} only when $P_s$ is present and enables the collaborative inference; otherwise, the same model should retain ordinary standalone reasoning.
We therefore expand the data into two prompt conditions.
\textbf{Corpus A} omits $P_s$, and none of its assistant trajectories contains \offload{}, preserving the non-collaborative behavior of $M_s$.
\textbf{Corpus B} augments the system message with $P_s$.
Its easy targets still omit \offload{} because making assistance available should not cause an unnecessary LLM call, while its hard targets contain between one and four dynamically inserted \offload{} tokens within the CoT segment to expose the model to possible token-level handoff locations.

These synthetic locations are not assumed to be the final optimal capability boundaries.
They provide Stage~2 with a behavioral cold start that distinguishes SLM-only completion from prompt-conditional offloading.
Stage~3 subsequently evaluates complete collaborative trajectories and adjusts the policy according to whether a handoff improves answer accuracy enough to justify its normalized cost.
The construction therefore supplies the initial action semantics without replacing the end-to-end accuracy--cost objective that ultimately determines when offloading is beneficial.

\subsubsection{Stage 1: Control-Token Embedding Learning}
\label{subsec:stage1-embedding-training}

The constructed trajectories specify where the handoff signal may occur, but the vocabulary of the original $M_s$ does not contain the special control token $\tau_{\mathrm{off}}$.
Stage~1 therefore expands the vocabulary and constructs a learnable representation for the new token before the model is trained to do offloading.
To avoid the unstable prediction behavior that can arise from fully random initialization, we initialize the input and output embedding vectors of $\tau_{\mathrm{off}}$ using the mean of a set of anchor embeddings and then add Gaussian noise:
\begin{equation}
  \mathbf{e}_{\tau_{\mathrm{off}}} = \frac{1}{|\mathcal{A}|} \sum_{a \in \mathcal{A}} \mathbf{e}_a + \epsilon, \quad \epsilon \sim \mathcal{N}(0,\, \sigma^2 \mathbf{I}),
  \label{eq:offload-embed-init}
\end{equation}
where $\mathcal{A}$ is a predefined set of anchor tokens representing natural breakpoints, including periods, newline characters, and end-of-sequence tokens, and $\mathbf{e}_a$ is the original embedding vector of the corresponding anchor token.
We set the noise standard deviation to $\sigma = 0.1$.
This scheme places $\tau_{\mathrm{off}}$ near common token boundaries in the embedding space, providing a stable initialization for Stage~1.

We then perform Stage~1 training on Corpus B constructed in Section~\ref{subsec:dataset-construction}.
During this stage, the embedding layer and output head are fully fine-tuned, while lightweight LoRA adapters attached to the backbone are updated jointly.
These transient adapters allow the backbone to adapt to the semantic embedding of the new control token while avoiding irreversible perturbations to the original base-model weights.
After training, we discard the transient LoRA adapters and merge only the trained parameter rows corresponding to $\tau_{\mathrm{off}}$ into the base model.
The resulting initialization of $M_s$ preserves the capabilities of the base model while enabling it to represent and recognize the control token.
At this point, \offload{} is representable by the model, but its use has not yet been conditioned on query difficulty or the presence of $P_s$; Stage~2 learns this behavioral distinction.

\subsubsection{Stage 2: Offloading-Oriented SFT for Behavioral Cold Start}
\label{subsec:stage2-cold-start}

Stage~2 implements the offloading-oriented SFT used for behavioral cold start in Figure~\ref{fig:training-pipeline} and converts the newly represented token into an initial offloading policy.
Starting from the $M_s$ produced by Stage~1, we perform one round of supervised fine-tuning on a mixture of Corpora A and B so that SLM-only completion and conditional offloading are learned jointly rather than as separate model variants.
Let $\mathcal{D}_A$ and $\mathcal{D}_B$ denote the training-example distributions induced by Corpora A and B, respectively.
Their samples are written as $(x,z_A)\sim\mathcal{D}_A$ and $(x,z_B)\sim\mathcal{D}_B$, where $x$ is the input context preceding the assistant response.
The target $z_A$ is a standalone response generated without the offloading prompt, whereas $z_B$ is the response under $P_s$: it contains no \offload{} for an easy example and includes the inserted control token for a hard example.
Stage~2 minimizes the negative log-likelihood over both corpora:
\begin{equation}
    \mathcal{L}_{\mathrm{SFT}}(\theta)
      = - \left( \mathbb{E}_{(x,z_A) \sim \mathcal{D}_A}\!\left[\log \operatorname{prob}_\theta(z_A \mid x)\right] + \mathbb{E}_{(x,z_B) \sim \mathcal{D}_B}\!\left[\log \operatorname{prob}_\theta(z_B \mid x,P_s)\right] \right),
  \label{eq:sft-loss}
\end{equation}
where $\operatorname{prob}_\theta(z \mid c)$ denotes the probability that $M_s$ with parameters $\theta$ assigns to response sequence $z$ given input context $c$.
The first term reinforces standalone SLM reasoning, whereas the second teaches prompt-conditional offloading when $P_s$ is enabled.
During this stage, the embedding layer remains frozen and LoRA is used to update the attention layers.
The resulting $M_s$ can reproduce the supervised offloading behavior, providing a stable policy initialization for collaborative rollouts in Stage~3.

\subsubsection{Stage 3: Cost-Aware Policy Alignment with GRPO}
\label{subsec:grpo}

The supervised targets establish how SLM-only completion and handoff should appear, but they do not measure the final answer produced after the CE invokes $M_l$ or the cost incurred by that joint trajectory.
Stage~3 closes this gap by introducing Group Relative Policy Optimization (GRPO)~\citep{shao2024deepseekmath} and executing the same one-way handoff used at inference time during policy rollouts.
\begin{algorithm}[!t]
  \footnotesize
  \caption{Three-stage progressive training of the SLM offloading policy}
  \label{alg:three-stage-training}
  \KwIn{base SLM $M_s^{(0)}$; frozen LLM $M_l$; training distributions $\mathcal{D}_A,\mathcal{D}_B,\mathcal{D}$; prompts $P_s,P_l$; control token $\tau_{\mathrm{off}}$; cost weight $\lambda$}
  \KwOut{trained SLM $M_s^{\star}$ parameterizing the offloading policy $\pi_s$}
  \LinesNotNumbered
  \makeatletter
  \algocf@seteveryparnl{\relax}
  \makeatother

  \begin{minipage}{\dimexpr\linewidth-2\algomargin\relax}
  \hrule
  \vspace{0.35em}
  \begin{minipage}[t]{0.475\linewidth}
    \textbf{Stage 1: Control-Token Embedding Learning}\par
    \vspace{0.18em}\hrule\vspace{0.3em}
    \nlset{1} Extend the vocabulary of $M_s^{(0)}$ with $\tau_{\mathrm{off}}$\;
    \nlset{2} Initialize its input/output embeddings using Equation~\ref{eq:offload-embed-init}\;
    \nlset{3} Attach transient LoRA adapters\;
    \nlset{4} Update the embedding layer, output head, and adapters on $\mathcal{D}_B$\;
    \nlset{5} Discard the adapters; retain the learned rows for $\tau_{\mathrm{off}}$ in $M_s^{(1)}$\;
  \end{minipage}%
  \hfill\vrule width 0.35pt\hspace{1.5em}
  \begin{minipage}[t]{0.46\linewidth}
    \textbf{Stage 2: Offloading-Oriented SFT}\par
    \vspace{0.18em}\hrule\vspace{0.3em}
    \nlset{6} Initialize $M_s^{(2)}\gets M_s^{(1)}$\;
    \nlset{7} Freeze the embedding layer and attach LoRA adapters to the attention layers\;
    \nlset{8}\ForEach{minibatch from $\mathcal{D}_A\cup\mathcal{D}_B$}{
      \nlset{9} Minimize $\mathcal{L}_{\mathrm{SFT}}$ in Equation~\ref{eq:sft-loss}\;
    }
    \nlset{10} Learn SLM-only completion and offloading\;
  \end{minipage}

  \vspace{0.55em}
  \hrule
  \vspace{0.35em}
  \textbf{Stage 3: Cost-Aware Policy Alignment with GRPO}\par
  \vspace{0.18em}\hrule\vspace{0.3em}
  \nlset{11} Initialize $\pi_\theta\gets M_s^{(2)}$ and keep $M_l$ frozen\;
  \nlset{12} Precompute $\operatorname{Cost}_{\mathrm{base}}^i$ for each $(q_i,y_i)\in\mathcal{D}$ using Equation~\ref{eq:grpo-baseline}\;
  \nlset{13}\While{the policy has not converged}{
    \nlset{14}\ForEach{example $(q_i,y_i)$}{
      \nlset{15} Sample $G$ SLM trajectories under $P_s$\;
      \nlset{16} Execute Algorithm~\ref{alg:ce-decode} to obtain joint outputs $O_{ij}$\;
      \nlset{17} Set $R_{\mathrm{acc}}^{ij}=\mathrm{Acc}(O_{ij},y_i)$ and compute $R_{\mathrm{eff}}^{ij}$ and $R_{\mathrm{total}}^{ij}$ using Equations~\ref{eq:grpo-eff} and~\ref{eq:grpo-total}\;
      \nlset{18} Compute $\widehat{A}^{ij}$ using Equation~\ref{eq:grpo-advantage}\;
    }
    \nlset{19} Update $\theta$ by minimizing $\mathcal{L}_{\mathrm{GRPO}}$ in Equation~\ref{eq:grpo-loss}\;
  }
  \nlset{20} Set $M_s^{\star}\gets\pi_\theta$\;
  \nlset{21}\KwRet{$M_s^{\star}$}
  \end{minipage}
\end{algorithm}

This aligns the SLM policy $\pi_s$ with the collaborative reward rather than with the placement of synthetic control-token labels alone.
The policy is parameterized by $\theta$, and $\pi_\theta$ is used interchangeably with $\pi_s$ below.
In the training implementation, the accuracy and normalized-cost terms in Equation~\ref{eq:problem-formulation} correspond to $R_{\mathrm{acc}}^{ij}$ and $R_{\text{eff}}^{ij}$, respectively, for rollout $O_{ij}$ of example $(q_i,y_i)$.

\textbf{The Reward Function of GRPO.} To couple policy learning with deployment cost, we define a baseline-normalized reward.
For each example $(q_i,y_i) \sim \mathcal{D}$, we first compute the cost $\operatorname{Cost}_{\mathrm{base}}^i$ of LLM-only inference on $q_i$ using $M_l$.
This baseline serves as a reference for evaluating the cost efficiency of the SLM-internalized offloading policy:
\begin{equation}
  \operatorname{Cost}_{\mathrm{base}}^i
    = C_{\mathrm{pre}}\,T_{\mathrm{in}}^i
    + C_{\mathrm{dec}}\,T_{l,\mathrm{dec}}^i,
  \label{eq:grpo-baseline}
\end{equation}
where $T_{\mathrm{in}}^i$ is the input-context length of request $q_i$, $T_{l,\mathrm{dec}}^i$ is the number of autoregressive tokens produced by $M_l$ during decoding when it executes the request alone, and $C_{\mathrm{pre}}$ and $C_{\mathrm{dec}}$ are the respective unit prices for LLM prefill and decoding.
We then model the absolute collaborative cost for each request $q_i$.
For each request $q_i$, the policy $\pi_\theta$ samples $G$ joint reasoning trajectories $O_{ij}$ ($j=1,\ldots,G$) under the offloading prompt $P_s$.
Once $\tau_{\mathrm{off}}$ appears in a trajectory, the training environment replays the offloading path in Lines~3--8 of Algorithm~\ref{alg:ce-decode}: it packages the partial reasoning trace, terminates SLM decoding, invokes the frozen LLM, and constructs the joint trajectory.
Consequently, the absolute collaborative cost $\operatorname{Cost}_{\mathrm{act}}^{ij}$ of each trajectory is defined as
\begin{equation}
  \operatorname{Cost}_{\mathrm{act}}^{ij}
    = (C_s + C_{\mathrm{pre}})\,T_s^{ij}
    + C_{\mathrm{dec}}\,T_{l,\mathrm{dec}}^{ij},
  \label{eq:grpo-cost-actual}
\end{equation}
where $T_s^{ij}$ is the number of tokens generated by the SLM $M_s$ before $\tau_{\mathrm{off}}$ is emitted, $T_{l,\mathrm{dec}}^{ij}$ is the number of tokens generated by the LLM $M_l$ during decoding after the handoff, and $C_s$ is a unified per-token price for the SLM.
Because the entire output of $M_s$ before $\tau_{\mathrm{off}}$ becomes part of the prefill context for $M_l$, $T_s^{ij}$ contributes to both SLM computation and LLM prefill cost.
Equation~\ref{eq:grpo-cost-actual} is the compact accounting form used by the training reward.
During evaluation, SLM prefill, SLM decoding, LLM prefill, and LLM decoding are measured separately and converted into monetary cost using Equation~\ref{eq:cost} in Section~\ref{sec:evaluation}.
Thus, the normalized cost relative to the LLM-only baseline is
\begin{equation}
  R_{\mathrm{eff}}^{ij}
    = \frac{\operatorname{Cost}_{\mathrm{act}}^{ij}}
           {\operatorname{Cost}_{\mathrm{base}}^i}.
  \label{eq:grpo-eff}
\end{equation}
For each joint trajectory $O_{ij}$, the task-accuracy reward is $R_{\mathrm{acc}}^{ij}=\mathrm{Acc}(O_{ij},y_i)$; it is computed by extracting \verb|\boxed{}| answers and applying equivalence checks such as \texttt{math\_verify}.
Combining this reward with the normalized efficiency penalty yields the total reward
\begin{equation}
  R_{\mathrm{total}}^{ij} = R_{\mathrm{acc}}^{ij} - \lambda R_{\mathrm{eff}}^{ij},
  \label{eq:grpo-total}
\end{equation}
where $\lambda \geq 0$ is the efficiency weight; its value can be selected through a sweep on the validation set.
When $R_{\text{eff}}^{ij} < 1.0$, collaborative inference costs less than LLM-only inference and therefore incurs a smaller penalty in $R_{\text{total}}^{ij}$.
When $R_{\text{eff}}^{ij} > 1.0$, redundant SLM generation or ineffective offloading increases the bill, thus reducing the trajectory's relative advantage within the group in GRPO.

\textbf{The Loss Function of GRPO.}
To apply the accuracy--cost trade-off encoded by Equation~\ref{eq:grpo-total} to a policy update, GRPO compares the total rewards of the $G$ rollouts sampled for the same request.
We compute the group-relative advantage as
\begin{equation}
  \widehat{A}^{ij}
    = \frac{R_{\mathrm{total}}^{ij}-\mu_i}{\sigma_i+\delta},
  \qquad
  \mu_i = \frac{1}{G}\sum_{j=1}^{G}R_{\mathrm{total}}^{ij},
  \qquad
  \sigma_i = \sqrt{\frac{1}{G}\sum_{j=1}^{G}
    \left(R_{\mathrm{total}}^{ij}-\mu_i\right)^2},
  \label{eq:grpo-advantage}
\end{equation}
where $\delta>0$ ensures numerical stability.
This query-wise normalization removes the need for a critic and makes the update depend on relative accuracy--cost performance within each group.
Let $a_{ij,t}$ be the $t$-th token sampled from the SLM policy, $h_{ij,t}$ its conditioning history, and $T_{\pi}^{ij}$ the number of SLM policy tokens, including $\tau_{\mathrm{off}}$ when emitted.
With the importance ratio
$\rho_{ij,t}(\theta)=\pi_\theta(a_{ij,t}\mid h_{ij,t})/\pi_{\theta_{\mathrm{old}}}(a_{ij,t}\mid h_{ij,t})$,
the GRPO loss is
\begin{equation}
  \mathcal{L}_{\mathrm{GRPO}}(\theta)
    = -\mathbb{E}_{(q_i,y_i)\sim\mathcal{D}}\!\left[
      \frac{1}{G}\sum_{j=1}^{G}\frac{1}{T_{\pi}^{ij}}
      \sum_{t=1}^{T_{\pi}^{ij}}
      \min\!\left(
        \rho_{ij,t}(\theta)\widehat{A}^{ij},
        \operatorname{clip}\!\left(\rho_{ij,t}(\theta),1-\epsilon,1+\epsilon\right)
        \widehat{A}^{ij}
      \right)
    \right],
  \label{eq:grpo-loss}
\end{equation}
where $\epsilon>0$ limits the policy update.
Only SLM-generated tokens enter Equation~\ref{eq:grpo-loss}; the frozen LLM completion affects the update through $R_{\mathrm{acc}}^{ij}$ and $R_{\mathrm{eff}}^{ij}$ but remains outside the gradient path.

Algorithm~\ref{alg:three-stage-training} consolidates the three-stage training pipeline after dataset curation.
Stage~1 learns a stable representation of the control action, Stage~2 converts it into an initial prompt-conditional offloading policy, and Stage~3 evaluates complete collaborative rollouts and aligns the policy with the proposed accuracy--cost reward and GRPO loss.
The learned SLM state passes from one stage to the next, while only the SLM policy is optimized and both the CE and $M_l$ remain unchanged; varying $\lambda$ yields the tunable accuracy--cost control described in the Introduction.
The resulting $M_s$ can therefore be deployed directly in the architecture of Section~\ref{subsec:architecture}, completing the connection from the objective in Equation~\ref{eq:problem-formulation} to the collaborative inference protocol.

\section{Experiments}
\label{sec:evaluation}

The key component of PyroDash is the handoff policy: the SLM must learn when to transfer its partial trajectory to the LLM, and the evaluation must consider both answer accuracy and the inference cost incurred by the handoff.
Our evaluation therefore answers four questions:
\begin{enumerate}[label=(\roman*)]
\item \textbf{Q1}: Does token-level offloading improve the trade-off between accuracy and cost relative to standalone inference?
\item \textbf{Q2}: Does the GRPO stage in Stage~3 produce behavior beyond the supervised cold start?
\item \textbf{Q3}: Does the efficiency penalty $\lambda$ provide useful control over dependence on the LLM?
\item \textbf{Q4}: Can the reported dollar savings be traced to measured SLM and LLM token usage?
\end{enumerate}
We next describe the experimental setup and answer the above questions with our main results, ablation studies and cost decomposition.

\subsection{Experimental Setup}
\label{subsec:experimental-setup}

\subsubsection{Models and Configurations}
\label{subsubsec:models}

In the evaluation, we select Qwen3.5-4B~\citep{qwenteam2026qwen35} and GLM-5.2-FP8~\citep{zai2026glm52fp8} as the SLM $M_s$ and the LLM $M_l$ in PyroDash, respectively.
This model configuration represents a typical deployment setting for PyroDash: the SLM handles reasoning by default, while the LLM is reserved for requests that the SLM decides to hand off.
We train the SLM with the three-stage pipeline as introduced in Section~\ref{sec:method} and keep $M_l$ frozen throughout the process.
After the three-stage pipeline, the trained $M_s$ is inserted into the Collaborate Engine (CE) used during collaborative inference, so training-time and evaluation-time handoffs follow the same one-way execution path.
We use $M_l$ only to read \localctx{} and complete a response after the CE initiates a handoff.

\subsubsection{Benchmarks, Baselines, and Metrics}
\label{subsubsec:benchmarks}

\textbf{Benchmarks.} Adaptive offloading is meaningful only when the evaluation contains both requests that the SLM can plausibly solve and requests for which stronger assistance from the LLM may be beneficial.
An easy suite would reward a policy that never calls the LLM, whereas an excessively hard suite could cause the learned policy to always offload.
We therefore choose five mathematical reasoning benchmarks spanning different levels of difficulty: GSM8K~\citep{cobbe2021gsm8k} covers grade-school word problems, Minerva~\citep{lewkowycz2022minerva} evaluates quantitative reasoning in technical domains, and Olympiad-Bench~\citep{he2024olympiadbench}, AIME-2024~\citep{huggingfaceh4_2024_aime}, and AIME-2025~\citep{lin2025aime} emphasize competition-level problems.
We report pass@1 accuracy under greedy decoding for Minerva, GSM8K, and Olympiad-Bench, and avg@32 for harder AIME problem sets to reduce sampling variation.
The aggregate score is the unweighted mean across the five benchmarks so that larger datasets do not dominate the comparison.

\textbf{Baselines.} We compare PyroDash with standalone SLM, LLM, SLM + SFT, and two routing baselines:
\begin{itemize}
  \item The standalone SLM and LLM baselines execute each request separately. They define the two endpoints of the deployment spectrum: the SLM-only result measures what can be obtained without the LLM's assistance, while the LLM-only result provides the evaluation-time quality and inference cost.
  \item SLM+SFT evaluates the SLM after Stages~1--2, including control-token embedding learning and the offloading-oriented behavioral cold start, but without cost-aware policy alignment.
  \item RouteLLM~\citep{ong2024routellm} represents request-level learned routing. Its matrix-factorization router assigns the entire query to either $M_s$ or $M_l$ before decoding, and its threshold $\alpha$ controls the fraction sent to the LLM.
  \item GlimpRouter~\citep{zeng2026glimprouter} provides a finer, training-free comparison: it uses the entropy of the first token at each reasoning step as an uncertainty signal and may call the LLM repeatedly. We set $\tau$ to $0.9$, the temperature to $0.6$, top-$p$ to $0.95$, and use double newlines as step boundaries. Together, RouteLLM and GlimpRouter contrast PyroDash with both coarse request-level assignment and external step-level intervention.
\end{itemize}

\textbf{Metrics.} We report four metrics for each method:
\begin{itemize}
  \item Accuracy (Acc.) measures each method's performance on each benchmark. We report per-benchmark accuracy and the average accuracy across all five benchmarks.
  \item LLM Token Ratio measures the proportion of decoded tokens produced by the LLM among all tokens decoded by the SLM and the LLM across all benchmarks:
\begin{equation}
  \text{LLM Token Ratio}
  = \frac{\sum n^{\text{llm}}_{\text{out}}}
  {\sum n^{\text{slm}}_{\text{out}} + \sum n^{\text{llm}}_{\text{out}}}\,,
  \label{eq:llm-token-ratio}
\end{equation}
where $n^{\text{slm}}_{\text{out}}$ and $n^{\text{llm}}_{\text{out}}$ denote the total number of output tokens decoded by $M_s$ and $M_l$, respectively, across all benchmarks.
This ratio exposes reliance on LLM generation, but it does not include prefill tokens and is therefore not a substitute for monetary cost.
  \item Avg.\ LLM Calls reports the mean number of calls to $M_l$ per example across all benchmarks.
  \item Cost measures each method's estimated total monetary cost in US dollars across all benchmarks, using the listed provider prices.
\end{itemize}

\subsubsection{Implementation Details}
\label{subsubsec:implementation}

We use the EasyHard-24k dataset~\citep{pyromind2026easyhard24k}, which contains 24{,}061 examples, in Stages~1--2.
In Stage~3, we use DAPO-Math released with DAPO~\citep{yu2025dapo}.
Training is implemented with TRL~\citep{vonwerra2020trl} and conducted using 4 NVIDIA H100 GPUs on the PyroMind platform~\citep{pyromind2026console}.
For Stages~1--2, we train LoRA adapters with rank $r{=}8$ and scaling factor $\alpha{=}16$ for 4 epochs at a learning rate of $1{\times}10^{-5}$.
For Stage~3, we use LoRA with $r{=}16$ and $\alpha{=}32$ for 1 epoch at a learning rate of $5{\times}10^{-6}$.
Each prompt produces $G{=}8$ rollouts, which are then used to compute the group-relative advantage in GRPO.
Before GRPO begins, we execute every training prompt with $M_l$ alone in order to obtain $\operatorname{Cost}_{\mathrm{base}}(q)$, which is used as the baseline.

We host GLM-5.2-FP8 on 8 NVIDIA B200 GPUs and Qwen3.5-4B on 4 NVIDIA H100 GPUs, respectively, using vLLM~\citep{kwon2023vllm} on the PyroMind platform.
Thinking mode is enabled in both models with a per-model limit of $\texttt{max\_tokens}{=}8192$. For each model, we keep the decoding configuration fixed across invocations.
The control token \offload{} is registered as a stopping sequence for SLM sampling.
Once it appears, the CE packages \localctx{} and performs the single handoff.

\subsection{Main Results}
\label{subsec:main-results}

\begin{table}[!t]
  \centering
  \footnotesize
  \setlength{\tabcolsep}{3pt}
  \caption{Performance on five mathematical reasoning benchmarks. The comparison includes the Qwen3.5-4B base model, the SFT cold start, RouteLLM, GlimpRouter, and PyroDash at two values of $\lambda$, with GLM-5.2-FP8 as the LLM-only reference. AIME results use avg@32; all other results use pass@1 under greedy decoding. LLM Token Ratio is defined by Equation~\eqref{eq:llm-token-ratio}. The highest benchmark accuracies and the lowest LLM-usage values among collaborative methods are shown in bold.}
  \label{tab:main-results}
  \begin{tabular}{lccccccrr}
    \toprule
    Method & \shortstack{Minerva\\Acc. (\%)} & \shortstack{GSM8K\\Acc. (\%)} & \shortstack{Olympiad\\Acc. (\%)} & \shortstack{AIME25\\Acc. (\%)} & \shortstack{AIME24\\Acc. (\%)} & \shortstack{Avg.\\Acc. (\%)} & \shortstack{LLM Token\\Ratio (\%)} & \shortstack{Avg.\\LLM Calls} \\
    \midrule
    Qwen3.5-4B & 25.00 & 77.56 & 25.78 & 8.75 & 4.69 & 28.36 & -- & -- \\
    Qwen3.5-4B+SFT & 43.75 & 89.69 & 47.70 & 21.88 & 28.23 & 46.25 & -- & -- \\
    \shortstack[l]{RouteLLM ($\sim$75\% GLM-5.2-FP8)} & 38.97 & 87.26 & 57.04 & 37.71 & 42.71 & 52.74 & 77.37 & 0.808 \\
    GlimpRouter & 45.59 & 93.63 & 51.56 & 35.10 & 45.10 & 54.20 & 75.11 & 1.20 \\
    PyroDash ($\lambda{=}0.05$) & \textbf{46.69} & 96.13 & \textbf{64.89} & \textbf{48.75} & \textbf{63.75} & \textbf{64.04} & 95.34 & 0.975 \\
    PyroDash ($\lambda{=}0.6$) & 44.85 & 90.90 & 60.44 & 34.17 & 42.40 & 54.55 & \textbf{1.90} & \textbf{0.012} \\
    \midrule
    GLM-5.2-FP8 & 43.75 & \textbf{96.44} & 61.04 & 40.62 & 46.56 & 57.68 & 100.00 & 1.000 \\
    \bottomrule
  \end{tabular}
  \vspace{0.75em}
  \small
  \caption{Averaged accuracy, LLM token ratio, average number of LLM calls, and cost across the five benchmarks.}
  \label{tab:cost-summary}
  \begin{tabular}{lrrrr}
    \toprule
    Method & Avg.\ Acc.\ (\%) & LLM Token Ratio (\%) & Avg.\ LLM Calls & Cost (\$) \\
    \midrule
    Qwen3.5-4B & 28.36 & 0.00 & 0.000 & 2.26 \\
    Qwen3.5-4B(+SFT) & 46.25 & 0.00 & 0.000 & 1.32 \\
    \shortstack[l]{RouteLLM ($\sim$75\% GLM-5.2-FP8)} & 52.74 & 77.37 & 0.808 & 44.62 \\
    GlimpRouter ($\tau{=}0.9$) & 54.20 & 75.11 & 1.20 & 31.61 \\
    PyroDash ($\lambda{=}0.1$) & 55.29 & 8.19 & 0.058 & 4.71 \\
    PyroDash ($\lambda{=}0.6$) & 54.55 & 1.90 & 0.012 & 1.78 \\
    \textbf{PyroDash ($\lambda{=}0.05$)} & \textbf{64.04} & 95.34 & 0.975 & 39.29 \\
    GLM-5.2-FP8 & 57.68 & 100.00 & 1.000 & 49.36 \\
    \bottomrule
  \end{tabular}
\end{table}

\begin{figure}[!t]
  \centering
  \includegraphics[width=0.8\textwidth]{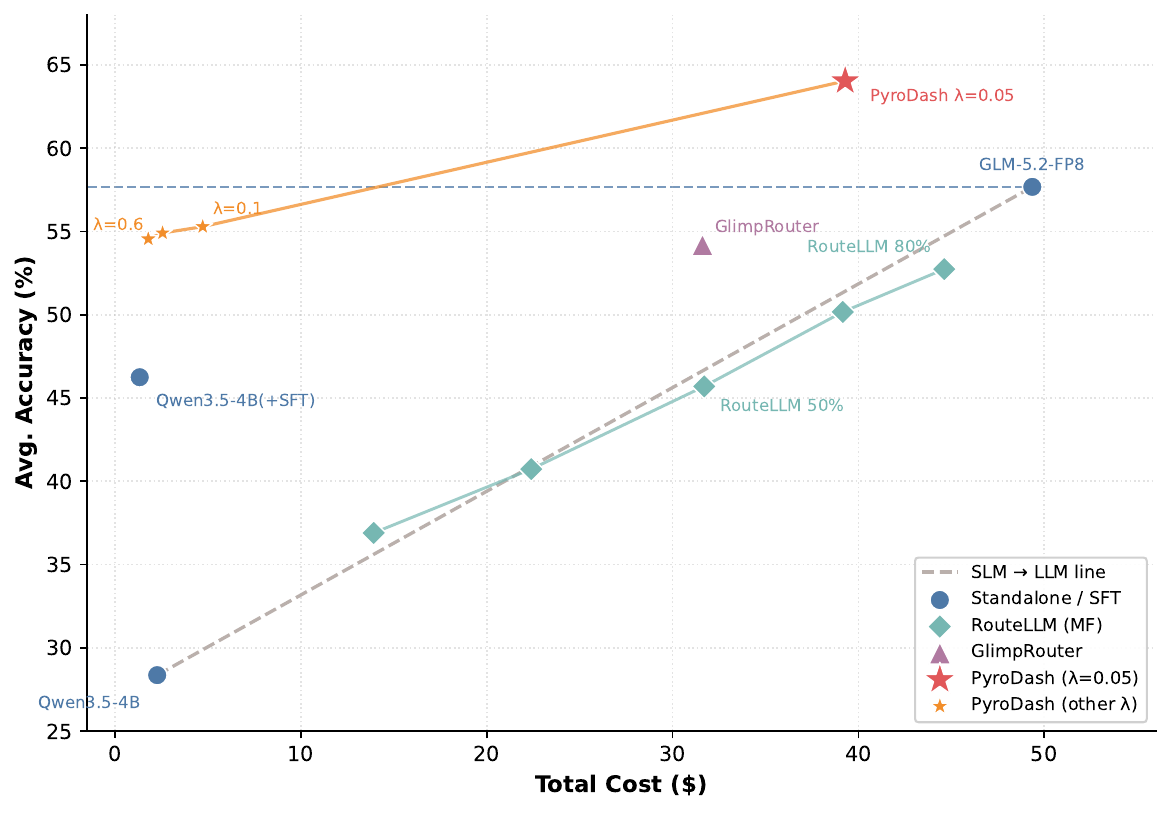}
  \caption{Total cost and average accuracy across five mathematical reasoning benchmarks. PyroDash with $\lambda{=}0.05$ reaches 64.04\% average accuracy, exceeding GLM-5.2-FP8's accuracy of 57.68\%, while PyroDash with $\lambda{=}0.6$ reduces total cost from \$49.36 to \$1.78, a reduction of 96.4\% relative to LLM-only inference.}
  \label{fig:cost-accuracy-pareto}
\end{figure}

We answer \textbf{Q1} by comparing PyroDash with standalone inference and two routing baselines in Table~\ref{tab:main-results} and Figure~\ref{fig:cost-accuracy-pareto}.
We report a quality-oriented policy at $\lambda{=}0.05$ and a cost-oriented policy at $\lambda{=}0.6$; the complete sweep is presented as the ablation study in Section~\ref{subsubsec:lambda-sensitivity}.

At $\lambda{=}0.05$, PyroDash reaches 64.04\% average accuracy, 6.36 percentage points above the accuracy of GLM-5.2-FP8, and obtains the highest reported accuracy on Minerva, Olympiad-Bench, AIME-2025, and AIME-2024.
At $\lambda{=}0.6$, PyroDash attains 54.55\% average accuracy with an LLM token ratio of 1.90\%, 0.012 LLM calls per example, and a total cost of \$1.78 as depicted in Table~\ref{tab:cost-summary}.
RouteLLM reaches 52.74\% at \$44.62, while GlimpRouter reaches 54.20\% at \$31.61; both allocate more than 75\% of decoded tokens to the LLM.
We conclude that PyroDash improves the balance between accuracy and cost: its quality-oriented policy outperforms the LLM-only baseline, while its cost-oriented policy matches or exceeds the routing baselines with substantially lower LLM usage and cost.

We answer \textbf{Q2} by comparing the SLM+SFT cold start with the fully trained PyroDash policy at $\lambda{=}0.05$ in Table~\ref{tab:main-results}.
SFT raises the standalone SLM average from 28.36\% to 46.25\%, whereas PyroDash at $\lambda{=}0.05$ reaches 64.04\%, including an increase from 28.23\% to 63.75\% on AIME-2024.
We conclude that the cost-aware policy alignment with GRPO in Stage~3 provides gains beyond the supervised cold start.

Figure~\ref{fig:cost-accuracy-pareto} and Table~\ref{tab:cost-summary} summarize the results for three representative values of $\lambda$.
The intermediate $\lambda{=}0.1$ policy reaches 55.29\% average accuracy at \$4.71, exceeding both routing baselines in accuracy at substantially lower cost.
Together, the policies trained with $\lambda{=}0.05$, $\lambda{=}0.1$, and $\lambda{=}0.6$ show that PyroDash supports distinct trade-offs between accuracy and cost.

\subsection{Qualitative Case Study}
\label{subsec:qualitative-case-study}

A short GSM8K example makes the one-shot execution path concrete.
The query asks when a lemon tree with a \$90 planting cost begins earning money if it produces 7 lemons per year, each sold for \$1.50, and incurs \$3 in annual maintenance.
The recorded PyroDash trajectory is:

\begin{pyrotracebox}{GSM8K \#12: from annual net income to first profit}
\pyroboxlabel{SLM PREFIX}
The SLM reduces the problem to an annual net-income calculation:
\[
7\times\$1.50-\$3.00=\$7.50.
\]
It then emits the learned control token.
\hfill\offload{}

\tcblower
\pyroboxlabel{LLM CONTINUATION}
After $n$ years, positive profit requires
\[
\$7.50n>\$90,
\]
so $n>12$.
Year 12 only recovers the planting cost, whereas year $\mathbf{13}$ is the first year with positive profit.
\end{pyrotracebox}

After detecting \offload{}, the CE removes it, sends the query and SLM prefix to the LLM for one continuation, and concatenates both segments.
This illustrates the inference flow only; four more cases, including an SLM-only completion, appear in Appendix~\ref{app:case-studies}.

\subsection{Ablation and Analysis}
\label{subsec:ablation-analysis}
\label{subsubsec:lambda-sensitivity}

\textbf{Sensitivity to the Efficiency Penalty $\lambda$.} We answer \textbf{Q3} by evaluating the $\lambda$ values reported in Table~\ref{tab:lambda-ablation} and Figure~\ref{fig:cost-accuracy-pareto}.
The sweep measures whether the efficiency penalty controls LLM dependence while preserving accuracy.
\begin{table}[H]
  \centering
  \small
  \caption{Sensitivity to the efficiency penalty coefficient $\lambda$.}
  \label{tab:lambda-ablation}
  \begin{tabular}{lrrrr}
    \toprule
    Configuration & Avg.\ Acc.\ (\%) & LLM Token Ratio (\%) & Avg.\ LLM Calls & Cost (\$) \\
    \midrule
    $\lambda{=}0.05$ & 64.04 & 95.34 & 0.975 & 39.29 \\
    $\lambda{=}0.1$ & 55.29 & 8.19 & 0.058 & 4.71 \\
    $\lambda{=}0.2$ & 54.91 & 3.18 & 0.026 & 2.55 \\
    $\lambda{=}0.3$ & 54.20 & 2.54 & 0.025 & 2.16 \\
    $\lambda{=}0.6$ & 54.55 & 1.90 & 0.012 & 1.78 \\
    \bottomrule
  \end{tabular}
\end{table}
From $\lambda{=}0.05$ to $0.1$, the LLM token ratio falls from 95.34\% to 8.19\%, average calls fall from 0.975 to 0.058, and cost falls from \$39.29 to \$4.71, while accuracy decreases from 64.04\% to 55.29\%.
For $\lambda \geq 0.1$, accuracy remains in the range of 54.20--55.29\% as the LLM token ratio decreases to 1.90\% and cost to \$1.78.
These results show that $\lambda$ provides useful control over dependence on the LLM, with the largest reduction in LLM usage observed when $\lambda$ increases from $0.05$ to $0.1$.

\textbf{Token Usage and Cost Decomposition.} We answer \textbf{Q4} by tracing total cost to measured SLM and LLM input and output tokens in Table~\ref{tab:cost-by-benchmark}.
We convert the four token components into dollar cost as
\begin{equation}
  C = n^{\text{slm}}_{\text{in}} p^{\text{slm}}_{\text{in}}
    + n^{\text{slm}}_{\text{out}} p^{\text{slm}}_{\text{out}}
    + n^{\text{llm}}_{\text{in}} p^{\text{llm}}_{\text{in}}
    + n^{\text{llm}}_{\text{out}} p^{\text{llm}}_{\text{out}},
  \label{eq:cost}
\end{equation}
where $n$ denotes the relevant token count and $p$ its unit price (in US dollars per million tokens).
We use $p^{\text{slm}}_{\text{in}}=\$0.05$/M and $p^{\text{slm}}_{\text{out}}=\$0.08$/M for Qwen3.5-4B, and $p^{\text{llm}}_{\text{in}}=\$0.90$/M and $p^{\text{llm}}_{\text{out}}=\$2.86$/M for GLM-5.2-FP8.
\begin{table}[H]
  \centering
  \small
  \setlength{\tabcolsep}{3.5pt}
  \caption{Token usage and cost of each benchmark. Input and output token counts are reported in millions and rounded to two decimal places before cost calculation. Note that discrepancies may arise from rounding.}
  \label{tab:cost-by-benchmark}
  \begin{tabular*}{\textwidth}{@{\extracolsep{\fill}} l *{15}{r} @{}}
    \toprule
    \multirow{3}{*}{\shortstack[l]{Benchmark}} &
    \multicolumn{5}{c}{Qwen3.5-4B} &
    \multicolumn{5}{c}{Qwen3.5-4B(+PyroDash, $\lambda{=}0.6$)} &
    \multicolumn{5}{c}{GLM-5.2-FP8} \\
    \cmidrule(lr){2-6} \cmidrule(lr){7-11} \cmidrule(lr){12-16}
    & \multicolumn{2}{c}{SLM} & \multicolumn{2}{c}{LLM} & &
    \multicolumn{2}{c}{SLM} & \multicolumn{2}{c}{LLM} & &
    \multicolumn{2}{c}{SLM} & \multicolumn{2}{c}{LLM} & \\
    \cmidrule(lr){2-3} \cmidrule(lr){4-5}
    \cmidrule(lr){7-8} \cmidrule(lr){9-10}
    \cmidrule(lr){12-13} \cmidrule(lr){14-15}
    & In & Out & In & Out & \$ &
    In & Out & In & Out & \$ &
    In & Out & In & Out & \$ \\
    \midrule
    Minerva &
    0.05 & 2.07 & 0.00 & 0.00 & 0.17 &
    0.05 & 0.22 & 0.00 & 0.00 & 0.02 &
    0.00 & 0.00 & 0.04 & 0.67 & 1.95 \\
    GSM8K &
    0.13 & 5.20 & 0.00 & 0.00 & 0.42 &
    0.17 & 0.32 & 0.00 & 0.00 & 0.03 &
    0.00 & 0.00 & 0.12 & 0.92 & 2.74 \\
    Olympiad &
    0.09 & 5.12 & 0.00 & 0.00 & 0.41 &
    0.11 & 2.10 & 0.01 & 0.02 & 0.24 &
    0.00 & 0.00 & 0.09 & 2.98 & 8.60 \\
    AIME25 &
    0.19 & 7.67 & 0.00 & 0.00 & 0.62 &
    0.22 & 4.77 & 0.05 & 0.06 & 0.61 &
    0.00 & 0.00 & 0.18 & 6.35 & 18.32 \\
    AIME24 &
    0.14 & 7.81 & 0.00 & 0.00 & 0.63 &
    0.16 & 4.66 & 0.07 & 0.15 & 0.87 &
    0.00 & 0.00 & 0.13 & 6.16 & 17.73 \\
    \midrule
    \textbf{Total} &
    \textbf{0.60} & \textbf{27.86} & \textbf{0.00} & \textbf{0.00} & \textbf{2.26} &
    \textbf{0.71} & \textbf{12.07} & \textbf{0.13} & \textbf{0.23} & \textbf{1.78} &
    \textbf{0.00} & \textbf{0.00} & \textbf{0.57} & \textbf{17.08} & \textbf{49.36} \\
    \bottomrule
  \end{tabular*}
\end{table}
At $\lambda{=}0.6$, PyroDash reduces LLM output tokens from 17.08M to 0.23M and total cost from \$49.36 to \$1.78 relative to LLM-only inference.
It produces 12.07M SLM output tokens, compared with 27.86M for standalone SLM inference.
We conclude that the reported savings are due to the reduced LLM decoding, while fewer total output tokens also contribute.

\section{Conclusion}
\label{sec:conclusion}

We propose PyroDash, a novel cost-aware, token-level paradigm for collaborative inference between an SLM and an LLM.
During autoregressive generation, the SLM emits the control token \offload{} when it requests stronger assistance.
The Collaborate Engine then packages the partial reasoning trace as \localctx{} and initiates a single handoff to a frozen LLM. This design requires neither LLM retraining nor a separate learned router.
The three-stage pipeline---control-token embedding learning, offloading-oriented supervised fine-tuning for behavioral cold start, and cost-aware GRPO alignment---trains $M_s$ under different values of $\lambda$ to support different trade-offs between accuracy and cost.
Unlike request-level routing, which assigns the entire query to either the SLM or the LLM before generation, PyroDash begins with the SLM and decides during generation whether to hand off the partial reasoning trace to the LLM.
The CE implements the handoff by sending the user query and the SLM's partial reasoning trace to the LLM as a new prompt, without requiring access to the LLM's hidden states or token probabilities.

Across five mathematical reasoning benchmarks, PyroDash with $\lambda{=}0.05$ achieves 64.04\% average accuracy, exceeding GLM-5.2-FP8's accuracy of 57.68\% while reducing total cost by 20.4\%.
The cost-oriented policy with $\lambda{=}0.6$ achieves 54.55\% average accuracy with an LLM token ratio of 1.90\% and 0.012 LLM calls per example.
Under the evaluated pricing model, it reduces total cost from \$49.36 to \$1.78, a reduction of 96.4\%.
RouteLLM and GlimpRouter allocate 77.37\% and 75.11\% of decoded tokens to the LLM, respectively.
PyroDash at $\lambda{=}0.6$ reduces this ratio to 1.90\%---about a 97.5\% relative reduction, or roughly one-fortieth of either baseline---while achieving higher average accuracy (54.55\% versus 52.74\% and 54.20\%).
The ablation study on $\lambda$ shows that a larger efficiency penalty leads to fewer LLM-generated tokens and lower inference cost.
Thus, PyroDash can be trained for either higher accuracy with greater LLM use or lower cost with less LLM use.

PyroDash shifts part of the scaling problem from enlarging a single model toward coordinating heterogeneous model services.
In this setting, an SLM handles the default queries, while a responsive LLM completes selected responses after the SLM initiates a handoff when the SLM detects the need for stronger assistance.
This collaborative inference paradigm allows PyroDash to balance reasoning accuracy against inference cost while remaining compatible with proprietary LLM APIs.
It also suggests a broader collaborative-scaling direction in which compute and intelligence can be collectively allocated across a mesh of model services rather than remaining tied to one set of weights.

\section{Limitations and Future Work}
\label{sec:limitations-future-work}

The current evaluation reports accuracy and LLM usage but does not examine the rationality of individual offloading decisions.
It therefore remains unclear whether the SLM hands off precisely when its reasoning becomes unreliable.
Future work should analyze trigger positions and relate them to specific reasoning failures.

Stage~3 penalizes inference cost relative to the cost of LLM-only inference.
Because we do not compare this normalized penalty with a direct token-cost penalty, the experiments do not isolate the benefit of normalization.
A controlled ablation of the two reward designs is therefore needed.

The present findings are limited to mathematical reasoning benchmarks, the EasyHard-24k data-curation procedure, and the Qwen3.5-4B and GLM-5.2-FP8 SLM-LLM model pair.
The performance of PyroDash in other application domains, such as code generation, tool use, and multimodal reasoning, as well as with other SLM--LLM combinations, remains to be explored.
Evaluating these settings will clarify how model capability and task structure affect learned offloading behavior.

The reported monetary costs are calculated from listed token prices rather than actual provider bills.
PyroDash allows only one handoff from the SLM to the LLM for each query without returning to the SLM after the handoff.
The experiments therefore do not show whether one handoff per turn would improve accuracy in multi-turn interactions.
Future work should compare the estimated costs with actual provider bills and evaluate a version of PyroDash that allows more than one handoff in multi-turn scenarios.

\bibliographystyle{plainnat}
\bibliography{references}

\clearpage
\appendix
\counterwithin{figure}{section}
\counterwithin{table}{section}
\counterwithin{equation}{section}
\counterwithin{algocf}{section}
\renewcommand*{\theHfigure}{appendix.\Alph{section}.\arabic{figure}}
\renewcommand*{\theHtable}{appendix.\Alph{section}.\arabic{table}}
\renewcommand*{\theHequation}{appendix.\Alph{section}.\arabic{equation}}
\providecommand*{\theHalgocf}{}
\renewcommand*{\theHalgocf}{appendix.\Alph{section}.\arabic{algocf}}

\section{Qualitative Case Studies of the PyroDash Inference Flow}
\label{app:case-studies}

This appendix complements the aggregate results in Section~\ref{sec:evaluation} with five illustrative cases drawn from the active PyroDash case set.
Four cases show how an SLM prefix is preserved across a one-shot handoff in algebra, materials science, arithmetic, and combinatorial reasoning, while one case shows how a tractable request remains entirely on the SLM.
The cases clarify the inference flow rather than provide an additional estimate of benchmark-level performance.

\subsection{Single-Handoff Completion after a Useful Partial Reduction}

\textbf{AIME-2024 \#744: solving a logarithmic system.}

\begin{pyropromptbox}
Let $x,y,$ and $z$ be positive real numbers satisfying
\[
\log_2\!\left(\frac{x}{yz}\right)=\frac12,\qquad
\log_2\!\left(\frac{y}{xz}\right)=\frac13,\qquad
\log_2\!\left(\frac{z}{xy}\right)=\frac14.
\]
If $\left|\log_2(x^4y^3z^2)\right|=\tfrac{m}{n}$ for relatively prime positive integers $m$ and $n$, find $m+n$.
\end{pyropromptbox}

The SLM first reduces the problem instead of handing it off immediately.
It then emits \offload{}, and the CE forwards the user prompt together with the partial trajectory to the LLM.

\begin{pyrotracebox}{One-shot handoff trace}
\pyroboxlabel{SLM PREFIX}
Let $a=\log_2x$, $b=\log_2y$, and $c=\log_2z$.
The constraints become
\[
a-b-c=\frac12,\qquad b-a-c=\frac13,\qquad c-a-b=\frac14,
\]
and the target is $\lvert4a+3b+2c\rvert$.
\hfill\offload{}

\tcblower
\pyroboxlabel{LLM CONTINUATION}
Adding the three equations gives $a+b+c=-\tfrac{13}{12}$.
Solving the system yields
\[
a=-\frac{7}{24},\qquad b=-\frac38,\qquad c=-\frac{5}{12}.
\]
Therefore, $\lvert4a+3b+2c\rvert=\tfrac{25}{8}$ and $m+n=\mathbf{33}$.
\end{pyrotracebox}

\textbf{Takeaway.}
The joint answer is correct, and the trace shows that the handoff preserves useful SLM work: the LLM continues from a reduced algebraic system rather than restarting from the natural-language question alone.

\clearpage
\subsection{Formula Selection before Numerical Completion}

\textbf{Minerva \#196: recovering atomic weight from a BCC unit cell.}

\begin{pyropromptbox}
A metal has a body-centered cubic (BCC) structure, a lattice constant of $3.31\,\text{\AA}$, and a density of $16.6\,\mathrm{g\,cm^{-3}}$.
Determine its atomic weight in grams per mole, rounded to one decimal place.
\end{pyropromptbox}

The SLM identifies the governing crystallographic relation, then hands off before performing the unit conversion and numerical calculation.

\begin{pyrotracebox}{One-shot handoff trace}
\pyroboxlabel{SLM PREFIX}
For a crystal with $n$ atoms per unit cell, atomic weight $A$, unit-cell volume $V_c$, and Avogadro constant $N_A$,
\[
\rho=\frac{nA}{V_cN_A}.
\]
The requested atomic weight can be obtained by rearranging this relation.
\hfill\offload{}

\tcblower
\pyroboxlabel{LLM CONTINUATION}
For BCC, $n=2$, and
\[
a=3.31\times10^{-8}\,\mathrm{cm},
\qquad
V_c=a^3=36.264691\times10^{-24}\,\mathrm{cm^3}.
\]
Therefore,
\[
A=\frac{\rho a^3N_A}{n}
=\frac{(16.6)(36.264691\times10^{-24})(6.022\times10^{23})}{2}
\approx\mathbf{181.3}\,\mathrm{g\,mol^{-1}}.
\]
\end{pyrotracebox}

\textbf{Takeaway.}
The trace separates model roles cleanly: the SLM supplies the governing equation, while the LLM completes the BCC-specific substitution, unit conversion, and rounding.

\subsection{Handoff after Reducing a Word Problem to an Inequality}

\textbf{GSM8K \#12: distinguishing break-even from positive profit.}

\begin{pyropromptbox}
Carlos pays \$90 to plant a lemon tree.
Each year it produces 7 lemons that he can sell for \$1.50 each, while watering and feeding cost \$3 per year.
After how many years will he start earning money from the tree?
\end{pyropromptbox}

The SLM converts the verbal description into an annual net-income calculation before requesting the LLM continuation.

\begin{pyrotracebox}{One-shot handoff trace}
\pyroboxlabel{SLM PREFIX}
Annual revenue is $7\times\$1.50=\$10.50$, and annual maintenance costs \$3.
The annual net income is therefore
\[
\$10.50-\$3.00=\$7.50.
\]
\hfill\offload{}

\tcblower
\pyroboxlabel{LLM CONTINUATION}
After $n$ years, cumulative net income is $\$7.50n$.
Positive profit requires
\[
7.50n>90,
\]
so $n>12$.
The tree breaks even at year 12 and first earns money in year $\mathbf{13}$.
\end{pyrotracebox}

\textbf{Takeaway.}
The SLM prefix preserves the useful arithmetic reduction, and the continuation resolves the strict inequality that distinguishes merely recovering the initial cost from earning a profit.

\clearpage
\subsection{Handoff after Establishing Combinatorial Constraints}

\textbf{AIME-2024 \#142: reconstructing a list from its sum, mode, and median.}

\begin{pyropromptbox}
A list of positive integers satisfies three conditions:
\begin{itemize}[leftmargin=1.5em,nosep]
  \item the sum of its entries is $30$;
  \item its unique mode is $9$; and
  \item its median is a positive integer that does not appear in the list.
\end{itemize}
Find the sum of the squares of all entries in the list.
\end{pyropromptbox}

The SLM identifies the structural consequence of the median constraint and corrects an initial false start before handing off the remaining case analysis.

\begin{pyrotracebox}{One-shot handoff trace}
\pyroboxlabel{SLM PREFIX}
The list must have even length; otherwise, its median would be an entry of the list.
For middle entries $a<b$, the median is
\[
m=\frac{a+b}{2},
\]
which must be an integer absent from the list.
The prefix briefly considers equal middle entries, rejects the resulting contradiction, and retains the distinct-middle-entry condition.
\hfill\offload{}

\tcblower
\pyroboxlabel{LLM CONTINUATION}
The unique mode condition requires at least two copies of $9$.
With four entries and exactly two $9$s, the other two entries must be distinct and sum to $12$.
Among the possible pairs, only $(5,7)$ gives an integer median absent from the list:
\[
\{5,7,9,9\},
\qquad
\operatorname{median}=8.
\]
The remaining feasible lengths fail the median or unique-mode condition.
Thus,
\[
5^2+7^2+9^2+9^2=\mathbf{236}.
\]
\end{pyrotracebox}

\textbf{Takeaway.}
The handoff occurs after the SLM isolates the list-length and middle-entry constraints but before exhaustive enumeration; the LLM completes the finite search and verifies the resulting list.

\subsection{SLM-Only Completion without an LLM Call}

\textbf{GSM8K \#217: retaining an easy request on the SLM.}

\begin{pyropromptbox}
There are 4 roses in a vase and 7 more dahlias than roses.
How many flowers are in the vase in total?
\end{pyropromptbox}

\begin{pyrotracebox}{SLM-only inference trace}
\pyroboxlabel{SLM COMPLETION}
The number of dahlias is $4+7=11$.
The total number of flowers is therefore
\[
4+11=\mathbf{15}.
\]
No handoff token is emitted, so the CE streams $O_s$ directly to the user and never invokes $M_l$.
\end{pyrotracebox}

\textbf{Takeaway.}
The CE follows Lines~9--11 of Algorithm~\ref{alg:ce-decode} and returns the SLM output directly.
Across the above cases, the trained SLM with GRPO either completes the request locally or marks a token-level boundary that triggers one context-preserving LLM continuation.

\end{document}